\documentclass[twocolumn]{article}
\usepackage{authblk}
\usepackage{geometry}
\newgeometry{top=3cm, bottom=2.5cm, left=2.5cm, right=2.5cm}
\usepackage{fancyhdr}
\usepackage{sectsty}
\sectionfont{\fontsize{12}{15}\selectfont}

\usepackage{microtype}
\usepackage{graphicx}
\usepackage{booktabs} 
\usepackage{nicematrix}
\usepackage[utf8]{inputenc}
\usepackage[english]{babel}
\usepackage{makecell}

\usepackage{wrapfig}
\usepackage{amsthm}
\theoremstyle{plain}
\theoremstyle{definition}
\usepackage{scalerel}
\usepackage{amsmath}
\usepackage{amssymb}

\usepackage{bm}
\usepackage{enumitem}
\setlist[enumerate]{label*=\arabic*.}
\usepackage{csquotes}
\usepackage{complexity}
\usepackage{xcolor}
\usepackage{subcaption} 
\usepackage{caption}
\usepackage{bbm}
\usepackage{amsbsy}
\usepackage{aligned-overset}
\usepackage[backend=biber, style=ieee]{biblatex} 
\usepackage{tikz}
\newcommand*\emptycirc[1][0.7ex]{\tikz\draw[thick] (0,0) circle (#1);} 
\newcommand*\halfcirc[1][0.7ex]{%
  \begin{tikzpicture}
  \draw[fill] (0,0)-- (90:#1) arc (90:270:#1) -- cycle ;
  \draw[thick] (0,0) circle (#1);
  \end{tikzpicture}}
\newcommand*\fullcirc[1][0.71ex]{\tikz\fill (0,0) circle (#1);} 
\usepackage{adjustbox}
\usepackage{caption}
\bibliography{arxiv_rl_scheduling_standardization}

\begin{document}

\title{\textbf{Towards Standardising Reinforcement Learning Approaches for Production Scheduling Problems}}

\author[1]{Alexandru Rinciog}
\author[2]{Anne Meyer}
\affil[1]{\small Faculty for Mechanical Engineering, TU Dortmund University. \texttt{alexandru.rinciog@tu-dortmund.de}}
\affil[2]{\small Faculty for Mechanical Engineering, TU Dortmund University. \texttt{anne2.meyer@tu-dortmund.de}}

\date{}

\twocolumn[
  \begin{@twocolumnfalse}
    \maketitle
    \begin{abstract}


Recent years have seen a rise in interest in terms of using machine learning, particularly reinforcement learning (RL), for production scheduling problems of varying degrees of complexity. 
The general approach is to break down the scheduling problem into a Markov Decision Process (MDP), whereupon a simulation implementing the MDP is used to train an RL agent. 
Since existing studies rely on (sometimes) complex simulations for which the code is unavailable, the experiments presented are hard, or, in the case of stochastic environments, impossible to reproduce accurately. 
Furthermore, there is a vast array of RL designs to choose from. To make RL methods widely applicable in production scheduling and work out their strength for the industry, the standardisation of model descriptions - both production setup and RL design - and validation scheme are a prerequisite. 
Our contribution is threefold: First, we standardize the description of production setups used in RL studies based on established nomenclature. 
Secondly, we classify RL design choices from existing publications. 
Lastly, we propose recommendations for a validation scheme focusing on reproducibility and sufficient benchmarking.
\end{abstract}

  \end{@twocolumnfalse}
\vspace{2em}
]


\section{Introduction}
\label{sec:introduction}
The field of production planning and scheduling is often characterized by a high degree of dynamism, given that production reality changes constantly. As production data and computing resources become more readily available and production itself becomes more versatile, the incorporation of machine learning techniques for production scheduling is increasingly an option. 


Production scheduling is NP complete for most real world cases, and, as such, difficult to solve optimally for large production instances.\cite{pinedo2012scheduling} Furthermore, unexpected events such as new job arrivals, operation duration deviations or resource availability issues, may invalidate schedules. In such cases fixed schedules need to be recomputed. 

Using production simulations that include unforeseen events alongside the scheduling constraints, RL agents can be trained to optimize a target goal by taking actions in response to changing production states. 
The RL solutions' adaptability and speed during deployment makes them attractive for high dimensional and/or stochastic problems \cite{waschneck2018optimization}. 

However, most studies suffer from reproducibility issues owing to a lack of clarity pertaining to RL modeling and/or production problem description, code and simulation input unavailability, and non-reproducible stochasticity. Furthermore, RL solutions should be benchmarked against more established approaches 
(e.g. local search \cite{vaessens1996job}) 
and typical simple heuristics on a sufficient number of scheduling problem instances to avoid cherry picking \cite{morse2010cherry} suspicion. A standard for discussing scheduling setups, RL solutions and RL results could alleviate the problems in the field.

This work seeks to standardise the description and validation of RL approaches for production scheduling problems. After confirming the standardisation gap in Section \ref{sec:related_work}, we move towards our goal in three steps: First, we look at production setups used with RL experiments and propose a clear nomenclature scheme by extending a well established Operations Research (OR) literature standard in Section \ref{sec:setup}. Secondly, in Section \ref{sec:rl_modeling}, we investigate RL design choices available for production, isolating different possible Markov Decision Process (MDP) formulations. Thirdly, we lay down how to ensure reproducibility and avoid cherry-picking in stochastic settings in section \ref{sec:val}. Section \ref{sec:conclusion} concludes our analysis and presents some avenues for future work.

As a basis for the categories introduced in this paper, we reviewed 37 publications containing experimental results on the  topic of RL production and scheduling.
The papers considered were published between 1998 and 2020. This material was compiled using the Web of Knowledge portal with \enquote{shop-scheduling} and \enquote{reinforcement learning} as search topics. 
The need for our work is reinforced by the significant number of more recent publications on the topic \cite{kayhan2021reinforcement}.

To keep within the frame of this work, we omit some of the details of the introduced categories. We do not 
look at all possible information available for state, action and reward formulations during RL design. 
Similarly, we do not consider RL algorithms in detail. Instead, we briefly describe the main categories and give examples of where these were used for production scheduling. 

\section{Related Standardisation Efforts}\label{sec:related_work}
A general standard for experiments in the field of OR is available in \cite{bartz2020benchmarking}. Here eight criteria for benchmarking are given, namely clearly stated goals, well-specified problems, suitable algorithms, adequate performance measures, thoughtful analysis, effective and efficient (experiment) designs, comprehensible presentations, and guaranteed reproducibility. 

For basic scheduling problems, e.g. flow/job/open shop scheduling problems, an implicit problem definition standard is set through the benchmarking instances available in Beasley's OR library \cite{beasley1990or}. New benchmarking instances are sometimes published alongside corresponding experiments, e.g. 
\cite{barnes1996flexible, dauzere1998multi}.

First steps in the direction of RL project standardisation were taken by OpenAI Gym, whereby a general RL application programming interface (API) is defined. Through \cite{hubbs2020or}, OpenAI Gym simulations for varied combinatorial optimization problems from the field of OR, e.g. the bin packing or the traveling salesman problem, are made available.

These standardisation efforts fail to address the particularities of the RL production scheduling field. Firstly, the RL literature defines many scheduling setups using inconsistent nomenclature. Secondly, RL offers many design choices for production scheduling problems. Finally, in stochastic settings, validation is difficult because of the high instance variance. 

\section{Production Scheduling Variants}
\label{sec:setup}
An established format for defining a production scheduling problem in OR literature is the $\alpha|\beta|\gamma$ notation described by Pinedo in \cite{pinedo2012scheduling}. The three parameters describe the machine setup $\alpha$, a possibly empty set of additional constraints $\beta$, and the optimization target $\gamma$. In what follows, we introduce the values these parameters can take and the relationship between them. We define new Pinedo-style values as needed to cover the setups present in RL scheduling literature relating them back to the values in  \cite{pinedo2012scheduling}. Figures \ref{fig:setup_hierarchy}, \ref{fig:beta} and \ref{fig:gamma} give an overview of the categories missing from \cite{pinedo2012scheduling} but described in RL literature and their relationship with the well-defined parameter values.

\subsection[Machine Setup]{Machine Setup -- $\bm{\alpha}$}
\label{subsec:alpha}
\begin{figure}[ht]
\begin{center}
\includegraphics[width=0.48\textwidth]{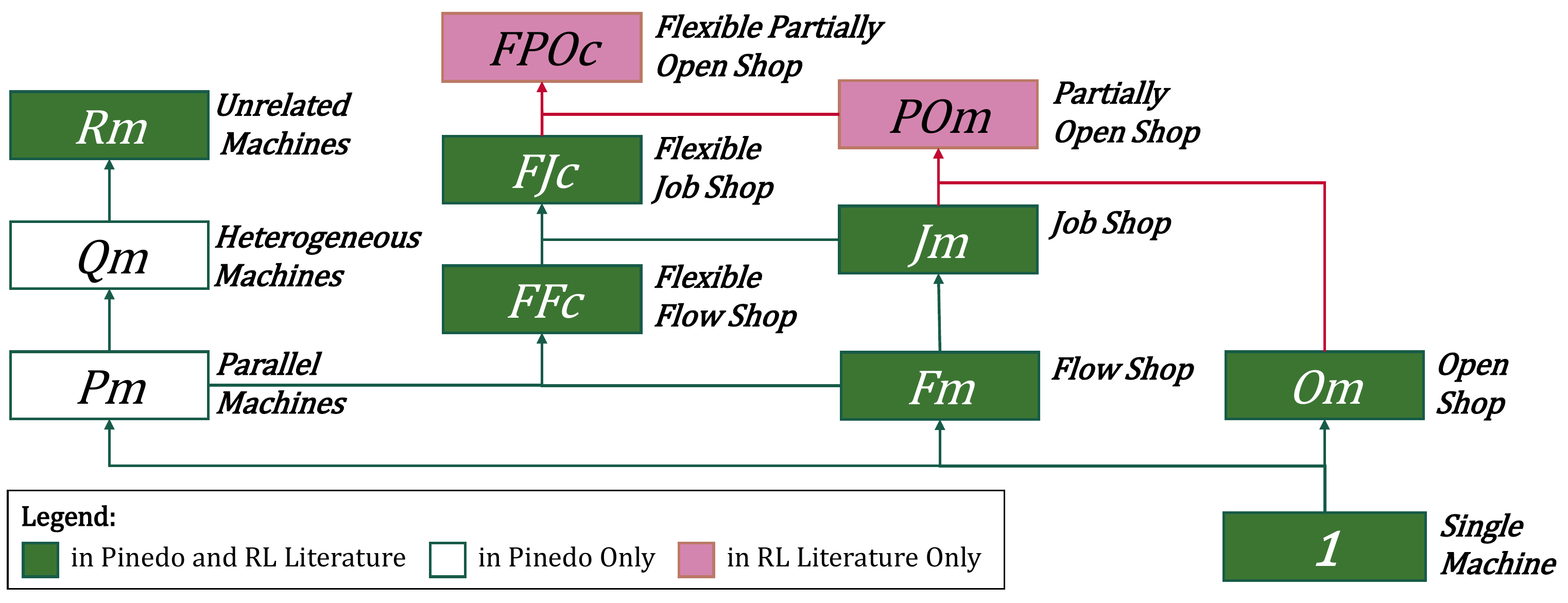}
\caption{Hierarchy of machine setups ($\alpha$).
}
\label{fig:setup_hierarchy}
\end{center}
\end{figure}

Production scheduling is the problem of assigning start times $s_{ji} > 0$ to a number of operations $o_{ji}$ with processing times $p_{ji}$ grouped into $n$ jobs, onto one of $m$ production resources to optimize some target value, for example the makespan.
The makespan  $C_\textbf{max}$ is defined as the maximum over the set of all job completion times $C_j$, with $j$ being the job index. Operations are executed one at a time in the absence of preemption. Release dates $r_j$ defining when job processing can commence and due dates (or deadlines) $d_j$ complete the picture. 
Depending on the number and order of operations within jobs and the number and speed of resources available for different operation types, machine setups can be categorized into one of the ten hierarchically arranged groups depicted in Figure \ref{fig:setup_hierarchy}.

Single machine problems ($1$) are defined in terms of $n$ jobs with one operation each, and a single processing resource. By replicating the processing resource we obtain the parallel machines ($Pm$) setup. $Pm$s with different processing speed scalars $v_i$ define the heterogeneous machines ($Qm$) setup. Operation processing times are herein scaled by the machine speed to obtain the operation completion times. If the machine speed depends not only on the machine index $i$ but also on the job index $j$, the unrelated machines ($Rm$) setup is present, e.g. \cite{zhou2020deep}.

Setups with $m$ resources and $n$ jobs containing $m$ operations, define shop-scheduling problems. Operations can only be processed by resources of corresponding types, and each resource is visited exactly once to complete a job. If operations within a job need to be processed in a fixed sequence,
which is identical for all jobs, then the production setup is that of a flow shop ($Fm$), e.g. \cite{mahadevan1998optimizing}. Allowing distinct sequences for different jobs yields a job-shop environment ($Jm$), e.g. \cite{gabel2012distributed}. No operation precedence constraints are indicative of an open-shop ($Om$), e.g. \cite{kim1998genetic}. 

Any shop setup can be extended by replicating resources and grouping them into work-centers. Operations are now fixed to a work-center, rather than a machine. If the setup prior to replication was flow or job shop, post replication the setup is that of a flexible flow shop ($FFc$), e.g. \cite{qu2015centralized}, or flexible job shop ($FJc$), e.g. \cite{park2019reinforcement}, respectively.

The class of partially open job shop scheduling problems, $POm$, has an intermediary position between $Jm$ and $Om$. Here job operations are partially ordered, with precedence constraints described by a directed acyclic graph.  Analogously to the other flexible variants, in flexible $POm$ (i.e. $FPOc$), operations of a particular type can be executed on one of the machines of the corresponding work-center. Examples of RL approaches over such setups can be found in \cite{zhang1996high, zhang1995reinforcement, hofmann2020autonomous}.

A routing choice for jobs is required additionally to operation sequencing decisions in parallel machine setups ($Pm$, $Qm$, $Rm$) and flexible shops. The same applies for setups where the precedence constraints allow for more than one operation to be executed immediately after a particular one has finished (i.e. in $Om$, $POc$).
We refer to this as routing flexibility.
\subsection[Additional Constraints]{Additional Constraints -- $\bm{\beta}$}
\label{subsec:beta}
\begin{figure}[ht]
\begin{center}
\includegraphics[width=0.485\textwidth]{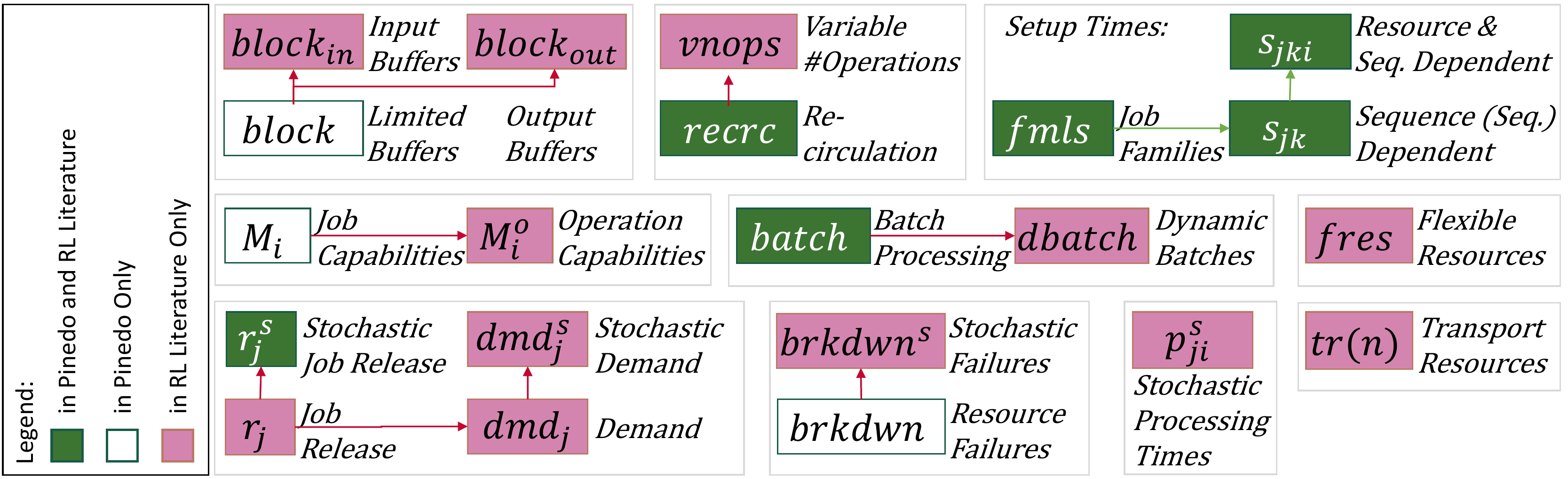}
\caption{Overview of additional constraints ($\beta$).
}
\label{fig:beta}
\end{center}
\end{figure}

The $\beta$ values modify the scheduling problem by adding constraints or changing existing ones. Figure \ref{fig:beta} contains an overview of the constraints to be discussed. We refer to \cite{pinedo2012scheduling} for some less frequent constraints (no wait-time  $nwt$, preemption $prmp$, permutation $prmu$, and \textit{job} precedence $prec$) that we do not extend in what follows. 

The $block$ parameter is defined by Pinedo to model limited buffer space between production stages in $Fm$. Machines can only process an operation if there is enough space in the buffer following it. We define the more generic parameters $block_{in}$, e.g. \cite{kuhnle2019design}, and $block_{out}$, e.g. \cite{mahadevan1998optimizing}, for input and output buffers in machine setups with routing flexibility. A full input buffer, blocks routing to the corresponding machine. Conversely, a full output buffer blocks processing on the afferent resource. 

The re-circulation parameter $recrc$, e.g. \cite{kuhnle2020designing}, loosens the constraint whereby jobs visit each machine type exactly once. Instead some jobs may require revisiting work-centers. This results in cycles in the operation precedence graph. We introduce the heterogeneous number of operations parameter, $vnops$, e.g. \cite{chen2010rule}, to model situations where jobs can revisit some machines while not visiting others at all, as is the case in \cite{zhang1996high, zhang1995reinforcement, rinciog2020sheet}.

Processing an operation often requires additional setup time. The parameters $fmls$, e.g. \cite{qu2015centralized}, $S_{jk}$, e.g. \cite{bouazza2017distributed}, and $S_{jki}$, e.g. \cite{hofmann2020autonomous}, define different setup time penalties in an increasing order of generality. With $fmls$ setup times are required when switching between job families. $S_{jk}$ penalizes switches between any jobs. 
For $S_{jki}$ setups, switch are additionally machine dependent.

Until now, any particular resource could only process one type of operations. This is not the case in setups of \cite{bouazza2017distributed, martinez2011reinforcement, jimenez2012generic,
luo2020dynamic}. Pinedo defines machine eligibility restrictions $M_i$ for $Pm$ environments only. Machine $i$ can only process operations from jobs contained in the set $M_i := J' \subset \mathcal{J}$, where $\mathcal{J}$ is the set of all jobs. 
We additionally generalize machine capability constraints for jobs to operation machine capabilities: 
To this end, we introduce the $M_i^o$ parameter containing a set of operation types where $M_i^o := O' \subset type(\mathcal{O})$ with $\mathcal{O}$ being the set of all operations and $type$ the mapping defining operation types.


The $batch$ parameter, e.g. \cite{qu2015centralized}, relaxes the constraint whereby resources can only process one operation at a time by allowing fixed size and processing time batches on specified resources. However, both batch sizes and processing time can be dynamic as in the setup of \cite{rinciog2020sheet}. We add the $dbatch$ parameter to reflect this. In the setup of  \cite{thomas2018minerva} resource capacity could vary dynamically, with resources being added to work-centers during the scheduling process. To cover this case, we define the flexible resources $fres$ parameter. The dynamic batch, and the flexible resource constraints are fringe cases in the scheduling literature, but potentially very impactful for real world applications. 

Parameters of core importance are those encoding stochasticity. 
Release dates $(r_j)$, could be deterministic, depending on the underlying planning process, but are most often stochastic ($r_j^s$), e.g. \cite{kuhnle2020designing}. Similarly, breakdowns are deterministic ($brkdwn$), if resources are taken offline for planned maintenance, and stochastic in case of unexpected failure ($brkdwn^s$), e.g. \cite{hofmann2020autonomous}. The presence of the demand parameter ($dmd_j$) implies that release dates are a system inherent choice. In a $dmd_j$ system, finished jobs are consumed from a sink buffer as per incoming demand. A stochastic demand is indicated by $dmd_j^s$. This situation is considered in \cite{mahadevan1998optimizing, qu2015centralized, paternina2005multi}. 
Operation processing times can also be seen as stochastic ($p_{ji}^s$), e.g. \cite{liu2020actor}, since these are often just estimates. 

Transport times and resources are often not explicitly modeled. In such a case, two implicit assumptions are made: (a) distances and hence transport times are either negligible, constant or part of the processing times $p_{ji}$, and (b) transport resources are always available to move jobs to their resource. As soon as (a) does not hold, transport times should be modeled separately. If the production setting to be scheduled contains enough transport resources, (b) can still hold. We mark this setup by means of $tr(\infty)$, e.g. \cite{thomas2018minerva}. If conversely, transport resources are scarce, we have to model their current position explicitly $tr(n)$ as in \cite{kuhnle2020designing, arviv2016collaborative}, since this transport availability directly impacts the schedule.
\subsection[Objectives]{Objectives - $\bm{\gamma}$}
\label{subsec:gamma}
\begin{figure}[ht]
\begin{center}
\includegraphics[width=0.48\textwidth]{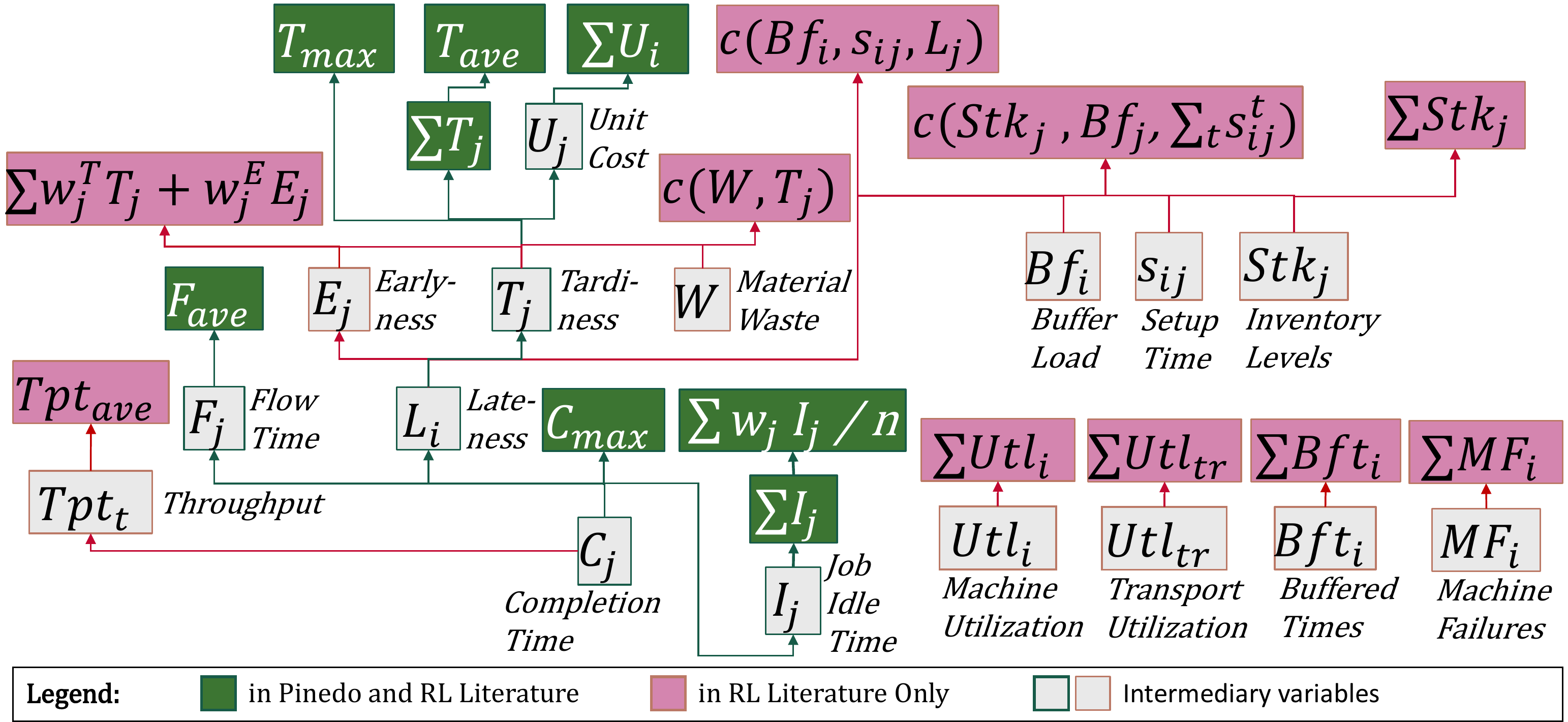}
\caption{Optimization targets in RL scheduling literature ($\gamma$).
}
\label{fig:gamma}
\end{center}
\end{figure}

As can be seen in Figure \ref{fig:gamma}, there are many possible optimization targets for production scheduling, for different use cases. We categorize optimization targets as job-centric (throughput, makespan, flow-time, lateness, tardiness, unit cost) or resource-centric (machine utilization, transport resource utilization, job idle time, machine failures, buffer length, buffered times, setup times, inventory levels). These metrics target different aspects of the production system and can be combined into a joint optimization target either by constructing a score function as a function of multiple targets (scalarization), e.g. \cite{qu2015centralized, rinciog2020sheet,  paternina2005multi, wang2020adaptive}, or by following a pareto front, e.g. \cite{mendez2019multi}.

\textbf{Job-Centric Goals:} The most often encountered metric (18 out of 37 publications) is the makespan ($C_{max}$), i.e. the maximum completion time over all jobs. While $C_{max}$ makes sense for static contexts with a fixed number of jobs, for dynamic environments a throughput measure needs to be used. In literature average job throughput $T\!pt_{ave}^j := 1/t \sum  1_{\{C_j \le  t\}}$ at time $t$ is used in  \cite{kuhnle2020designing, thomas2018minerva}. Additionally, operation throughput  $T\!pt^o_{ave}:= 1/t \sum_{j,i} 1_{\{s_{ji} + p_{ji} \le  t\}}$ could be considered. Flow-time (also lead time) $F_j$ measures the time between job release and job completion, $F_j := C_j - r_j$. The average flow time $F_{ave}:= 1/n \sum F_j$ is an indicator of a system's reactivity/flexibility. Yet another measure based on the job completion time is the job idle time $I_j$ defined as the difference between flow time and the summed job processing time: $F_j - \sum_i p_{ji}$. 

Perhaps the most relevant metrics for manufacturing companies are timeliness related \cite{schuh2013logistikmanagement}. Timeliness metrics are computed using the lateness intermediary variable, i.e. the difference between completion time and due date, $L_j := C_j - d_j$.
Tardiness extends lateness by ignoring early jobs $T_j := \text{max} \{0, L_j\}$. Tardiness based optimization goals encountered are constructed by aggregating the tardiness variables, e.g. $T_{ave}, T_{max}, \sum T_j$. Alternatively, the number of tardy jobs can simply be counted using the unit cost variables $U_j := 1_{\{C_j > d_j\}}$, e.g. \cite{wang2005application}. Finishing jobs too early can be detrimental \cite{baker2014minimizing}. As such, earliness $E_j:= \vert \text{min}\{L_j, 0\} \vert$ can be another minimization target as it is considered in \cite{wang2020adaptive}.

\textbf{Resource-Centric Goals:} Resource utilization is defined as $Utl_i := 1/t \sum_j p_{ji} \cdot 1_{\{s_{ji} + p_{ji} \le  t\}}$, i.e. the time the machine was working over the up-time $t$. Similarly one can define the transport utilization $Utl_{tr}$ as the time spent carrying a load over the total elapsed time. The buffer composition can also be used as an optimization target. In general one would like to keep buffer lengths to a minimum. In \cite{qu2015centralized} the number of buffered operations $Bf_i$ at resource $i$ was used as an optimization goal within a more complex cost function also involving lateness and the total number of tool switches. 


\section{RL Modeling for Production Scheduling}
\label{sec:rl_modeling}
To solve any problem with RL first an MDP must be modeled for the particular domain. The MDP defines the agent-environment interaction. The agent, whose goal is to maximize the received reward, senses the current environment state and takes an action, whereby the environment is moved to a new state. This loop continues until an end-state is reached. The agent receives feedback by means of a reward signal, which it tries to maximize. While for many RL settings the MDP is fairly obvious, for production scheduling environments, there are many design choices to be made. In what follows we categorize the RL design options available for production scheduling problems. 

\subsection{MDP Breakdown}
\label{subsec:mdp}
We start by defining five broad categories of MDPs (i.e. MDP breakdowns) present in RL scheduling literature:
operation sequencing (1), routing before sequencing (2), interlaced routing and sequencing (3) transport-centric routing (4), and re-scheduling (5). In addition, we propose the holistic routing and sequencing breakdown (6) to better tackle $tr(n)$ settings. 

\textbf{Operation sequencing:} The idea behind iterative operation sequencing is to use the moments when resources become free as the discrete time steps when decisions can be taken. An operation $o_{ji}$ from the machine queue is then assigned to it, marking its start time $s_{ji}$ as the current time $t$. This process is repeated so long as there are still jobs to complete. In setups with no routing flexibility, the iterative operation sequencing MDP breakdown is sufficient for defining all possible schedules, e.g. \cite{mahadevan1998optimizing, kim1998genetic, riedmiller1999neural, aydin2000dynamic, gabel2007successful}. 

Iterative operation sequencing can also be used to implicitly define routes for setups with routing flexibility, if the assumption of negligible transport times and endless capacity buffers is present. After an operation finishes, the next eligible resources for the job can be seen as sharing a virtual buffer wherein the job lies. As soon as the job is assigned by the agent to one of these resources, the job is no longer available for processing at the other resources. Such an approach can be found in \cite{waschneck2018optimization, zhou2020deep, qu2015centralized, rinciog2020sheet, luo2020dynamic, thomas2018minerva}.

\textbf{Routing Before Sequencing:} 
Within this breakdown, whenever a new job arrives, the agent iteratively assigns processing resources to it, until it has a fixed completion path through production. When the job route has been set, the scheduling process described above can continue. Such is the approach of \cite{martinez2011reinforcement, jimenez2012generic}.

\textbf{Interlaced Routing and Sequencing:}
The combined routing and sequencing problem can also be solved in an interlaced fashion. The agent decides on demand upon the next processing resource to transport a job to, when an operation is finished. This decision is then immediately followed by a sequencing decision on the freed resource. This breakdown is employed in \cite{bouazza2017distributed}.

\textbf{Transport-Centric Routing:}
In $tr(n)$ settings, there are two behaviorally distinct resource types present: processing and transport resources. Within this MDP breakdown, processing resources sequence operations following a priority rule, e.g. first-in-first-out (FIFO). Decisions are required when transport resources are idle. The agent alternates between the selection of source processing resources and destination processing resources. The oldest job in the source output buffer is chosen and transported to the destination input buffer. Examples are \cite{kuhnle2020designing,arviv2016collaborative, stricker2018reinforcement}. 

\textbf{Holistic Routing and Sequencing:}
The caveat of the breakdown above is that not all schedules are possible. To complete the picture we propose that decisions be requested of the agent either when transport or processing resources are idle. In the case of transport resources, the agent behavior is the same as in the transport-centric breakdown. When dealing with processing resources, the agent should behave as in the operation sequencing breakdown.

\textbf{Re-Scheduling:}
Finally, whenever a stochastic event occurs, the agents can be tasked with deploying an external solver to compute a new plan. Such is the case in \cite{shahrabi2017reinforcement}, where an agent selects the parameters of a variable neighborhood search (VNS) \cite{mladenovic1997variable} on new job arrivals. VNS is used to create a new schedule which is followed until a new job arrival.


\subsection{Agent Types}
\label{subsec:agents}

RL agents are trained according to the generalized policy iteration principle. Herein two stages are distinguished, namely policy evaluation and policy improvement. During policy evaluation, agents select actions according to a policy function, and observe their returned reward. During policy improvement, the policy is adjusted based on the observations made. These stages are repeated until convergence (hopefully).
RL algorithms can be classified along three discrete axes:  value, policy or actor-critic methods (1), model-free or model-based methods (2), and on- or off-policy methods (3) (cmp. \cite{sutton2018reinforcement}).  

\textbf{Value, Policy and Actor-Critic Methods:} Value methods try to estimate future reward by means of a value function, which is used to estimate the \enquote{goodness} of either states or actions from given states. During the policy evaluation stage, actions are selected by using the value function to ascertain the quality of the states reachable from the current one. The observed rewards are used during the subsequent stage to improve the value function. Examples of such methods are State Action Reward State Action (SARSA) \cite{rummery1994line}, Q-Learning (QL) \cite{watkins1989learning}, Deep Q-learning (DQN) \cite{mnih2013playing}, and Double DQN (DDQN) \cite{van2016deep}. QL is particularly popular with the production scheduling community with implementations using tables to represent value functions used in \cite{qu2015centralized, chen2010rule, bouazza2017distributed, wang2005application, shahrabi2017reinforcement}, or neural networks as function approximators in \cite{jimenez2012generic, paternina2005multi,  wang2020adaptive, aydin2000dynamic}. DQN was used in \cite{hofmann2020autonomous} and DDQN in \cite{ zhou2020deep, park2019reinforcement, luo2020dynamic}, 

Alternatively, agent policies $\pi_\theta$ can be used directly to select actions during the evaluation stage. 
Based on the observed rewards, the reward expectation under policy $\pi_\theta$ is estimated, its gradient with respect to $\theta$ is computed and the policy is updated using stochastic gradient ascent.  
Examples include REINFORCE \cite{williams1987class} used for production scheduling in \cite{gabel2012distributed}, Trust Region Policy Optimization (TRPO) introduced in \cite{schulman2015trust} and used for instance in \cite{kuhnle2020designing} and Proximal Policy Optimization (PPO) \cite{schulman2017proximal} used in \cite{zhang2020learning}, albeit for a deterministic scheduling problem.

Policy and value approaches can be combined into actor-critic algorithms. Instead of using environment interaction to approximate the expected reward directly, the (state) value function approximator (critic), is used to inform the policy approximator (actor) of the quality of its action. AlphaZero (AZ) \cite{silver2017shogi} used in \cite{rinciog2020sheet} and Deep Deterministic Policy Gradient (DDPG) \cite{lillicrap2015continuous} used in \cite{liu2020actor} fall in this category.

\textbf{On- vs Off-Policy:}
On policy methods (e.g. SARSA) use the same policy during evaluation stage that was adjusted during the improvement stage. This leads to more stable learning at the expense of exploration, which can lead to local optima. Conversely, in off-policy methods (e.g. QL, DQN, DDQN), the policy used during the evaluation stage can differ from the one used in the improvement stage, which leads to more exploration at the expense of convergence speed.

\textbf{Model Based vs. Model Free:} 
RL algorithms can be furthermore split into model-free and model based approaches. Model-based approaches use an environment model to plan a few steps into the future before deciding on an action. The involved environment model is either estimated by the agent itself, e.g. Imagination Augmented Agent (I2A) \cite{racaniere2017imagination}, or simply given to it, e.g. AZ. 

\textbf{Single- vs Multi-Agents:}
We can allow multiple agents to act within the same environment. These agents can be cooperative, i.e. striving to jointly maximize the expected reward or competitive, with each agent targeting a maximization of his reward only. For production scheduling multi-agent systems are often deployed (17 of the 37 studies). Depending on the MDP breakdown, agents can be associated with different setup components.
Agents posted with each processing resource for sequencing decisions can be found in  \cite{waschneck2018optimization, bouazza2017distributed, jimenez2012generic, liu2020actor, wang2020adaptive}. 
Agents deployed per job for sequencing are used in \cite{paternina2005multi} and for routing in \cite{jimenez2012generic}.
Routing agents are deployed per job family by \cite{bouazza2017distributed}) or per transport resource in \cite{arviv2016collaborative}. 


\subsection{State, Action and Reward Modeling}
\label{subsec:state}
\textbf{States}: The information  based on which the state transition occurs, i.e. the environment state, depends on the production setup considered and the MDP breakdown chosen. Take, for instance, a standard job-shop scheduling problem $Jm|C_{max}$ with a sequencing breakdown. Here we encode the complete state information as four $n \times m$ dimensional matrices, the system time $t$ and index $i$ of the processing resource requesting a scheduling decision: $(T, P, L, A, t, i)$. 
The type matrix $T$ keeps track of the precedence within jobs (rows) and the type of resource the job operations need. The corresponding values in the $P$ matrix represent the remaining processing time for each operation. The location of all operations is be encoded using a matrix $L := (l_{ji})$, with $l_{ji} = m_k$ if operation $o_{ji}$ is at the machine $m_k$. To distinguish between  operations being actively processed and those waiting in the machine buffer, the boolean matrix $A$ is used. 


If we present all the environment state information  to the agent together with a machine requesting a new operation, we are in a fully observable environment situation. 
This means that the agent could, in theory, perfectly approximate the state transition and reward functions. 
However, a large state space and limited computational resources can make learning very time consuming. 
Additionally, it could be desirable that the agent learns a good policy with only local information, to  enable the production system to easily scale up (e.g. by adding a new resource), without the need of a global monitoring system.

As such, the agent is mostly only presented with a subset of the environment state information available, which corresponds to a partially observable environment. We distinguish between raw state information, e.g. any subset of the $\{T, P, L, A, t, i\}$ in the situation above, or information in condensed form, i.e. environment state features, or both. Employed features fall into one of three categories, namely job properties, resource attributes and optimization targets. 

Job property features condense job information. Examples of such are remaining job operations, remaining job processing time \cite{gabel2009multi} or number of jobs in the system \cite{shahrabi2017reinforcement}. Resource related features target an aggregation of machine properties, such as remaining processing time in buffers \cite{zhou2020deep}, resource workload \cite{wang2020adaptive, gabel2009multi}, ratio of remaining processing times for machines and buffered processing time \cite{chen2010rule}, number of product types in each buffer or machine health \cite{qu2015centralized}. Estimates of the target function, or of other targetable indicators, are also often included in the construction of the agent state. Such examples are estimated total tardiness \cite{luo2020dynamic, wang2005application}, average machine utilization \cite{luo2020dynamic, thomas2018minerva}, average transport resource utilization or the aforementioned average buffer length \cite{thomas2018minerva}.

\textbf{Actions}: In terms of actions, there are two main approaches to production scheduling. The agent either takes an action directly, or selects an appropriate priority rule for scheduling or routing. 

In the operation sequencing MDP, direct actions are operation indices, e.g. \cite{gabel2009multi, fonseca2018q}, or job indices, e.g. \cite{jimenez2012generic}, depending on whether the job operations are totally ordered (latter) or not (former). For the routing decision, the raw actions are processing resource indices, e.g. \cite{kuhnle2020designing}. 

The basic intuition for the priority rule approach is that different rules will be more effective in different production scenarios. Given a state, the agent will output a rule from a fixed set. 
The rule is then applied to prioritize the operations in the buffer (iterative scheduling) or the viable downstream processing resources (iterative routing). The highest priority operation/machine gets selected. Mostly, the encountered priority rules are very simple. The most frequent examples of operation prioritization rules are shortest processing time (SPT), longest processing time (LPT), earliest due date (EDD), first in first out (FIFO) or last in first out (LIFO). The rule names are self explanatory. Machine prioritization rules work analogously (e.g. \cite{bouazza2017distributed}). One could, for instance, prioritize the downstream machine with the shortest queue (SQ) in terms of processing time, the one with the least elements (LQE) or the shortest setup times (SST).

A completely different approach to action modeling was taken in \cite{shahrabi2017reinforcement}. The RL agent gets deployed on occurrence of new job arrivals. At this point a VNS is started to recompute a schedule for the operations not yet started. The RL agent action outputs the VNS parameters for the given state.

\textbf{Reward:}
The last modeling decision to be made pertains to the reward signal produced by the environment. For production scheduling, the reward is most often proportional to the optimization target or a value that highly correlates with it (e.g. makespan and average utilization). That being said, there is no universally accepted scheme for constructing a reward function.
In \cite{jimenez2012generic, gabel2007successful, gabel2009multi} the number of queued operations at the current machine is used to build the reward signal (the shorter the queue the better), though the optimization target was makespan.

Reward functions also differ based on when the agent receives a non-zero reward, e.g., at every decision point, at the end of the game, or at any point in-between. Other design choices include whether the reward is discrete or continuous \cite{matignon2006reward}, strictly positive or both positive and negative and many more \cite{sutton2018reinforcement}. Finding an appropriate reward is of singular importance, given that it strongly determines learning convergence, and can be a challenging task for scheduling \cite{lang2020integration}.

\section{Validation}
\label{sec:val}
In this section investigate what is needed in order to ensure  reproducibility and provide sufficient validation of RL production scheduling experiments. Table \ref{tab:reproducibility&validation_static} summarizes the relevant stochastic RL literature in terms of production setup clarity and RL design clarity, simulation input availability, reproducible stochasticity, train-test split, state of the art coverage through baseline algorithms and, finally, cherry picking potential. 

\captionsetup{belowskip=5pt}

\begin{table}[ht]
\scriptsize
\centering
\caption{Reproducibility and baselining fulfillment for stochastic RL production scheduling; Filled, and empty circles for fulfilled and unfulfilled criteria. }
\setlength{\tabcolsep}{4pt} 
\begin{tabular}{rccccccc}
\cmidrule(l){1-8}
\textbf{Source}  &
  \multicolumn{4}{l}{\textbf{Reproducibility}} &
  \multicolumn{3}{l}{\textbf{Evaluation}} \\ 
\cmidrule(l){1-1}\cmidrule(l){2-5}\cmidrule(l){6-8}
      
 &
  \rotatebox{90}{\makecell[l]{Setup\\Clarity}} &
  \rotatebox{90}{\makecell[l]{RL\\Clarity}} &  
  \rotatebox{90}{\makecell[l]{Input\\Availability}} &
  \rotatebox{90}{\makecell[l]{Reproducible~~ \\Stochasticity~}} &
  \rotatebox{90}{\makecell[l]{Train-Test\\Split}} & 
  \rotatebox{90}{\makecell[l]{Sufficient\\Baselines}} &
  \rotatebox{90}{\makecell[l]{Cherry\\Picking~ \\Potential}} \\ 

\cmidrule(l){1-1}\cmidrule(l){2-5}\cmidrule(l){6-8}
\cmidrule(l){1-1}\cmidrule(l){2-5}\cmidrule(l){6-8}

\makecell[r]{\cite{waschneck2018optimization, zhou2020deep, bouazza2017distributed}}& 
	\halfcirc & \halfcirc &  \emptycirc & \emptycirc & 
	\emptycirc & \halfcirc & \halfcirc \\ 
\cite{qu2015centralized} & 
	\fullcirc & \fullcirc &  \emptycirc & \emptycirc & 
	\fullcirc & \halfcirc & \halfcirc \\		
\cite{park2019reinforcement} & 
	\halfcirc & \halfcirc & \emptycirc & \emptycirc & 
	\fullcirc & \fullcirc & \emptycirc \\
\cite{hofmann2020autonomous} & 
	\fullcirc & \fullcirc &  \emptycirc & \emptycirc & 
	\emptycirc & \halfcirc & \emptycirc \\	
\cmidrule(l){1-1}\cmidrule(l){2-5}\cmidrule(l){6-8}

\cite{kuhnle2019design} & 
	\halfcirc & \fullcirc & \emptycirc & \emptycirc & 
	\emptycirc & \emptycirc & \emptycirc \\	
\cite{kuhnle2020designing} & 
	\fullcirc & \fullcirc &  \emptycirc & \halfcirc & 
	\fullcirc & \halfcirc & \emptycirc \\ 	
\cite{chen2010rule} & 
	\fullcirc & \fullcirc &  \emptycirc & \emptycirc & 
	\fullcirc & \halfcirc & \emptycirc  \\
\cite{jimenez2012generic} & 
	\halfcirc & \halfcirc &  \halfcirc & \emptycirc & 
	\emptycirc & \halfcirc & \halfcirc \\
\cmidrule(l){1-1}\cmidrule(l){2-5}\cmidrule(l){6-8}

\cite{luo2020dynamic, wang2005application} & 
	\fullcirc & \fullcirc &  \emptycirc & \emptycirc & 
	\emptycirc & \halfcirc & \halfcirc \\
\makecell[r]{\cite{thomas2018minerva, aydin2000dynamic, shahrabi2017reinforcement}} & 
	\halfcirc & \fullcirc &  \emptycirc & \emptycirc & 
	\emptycirc & \halfcirc & \halfcirc \\
\cite{liu2020actor} & 
	\fullcirc & \fullcirc &  \fullcirc & \emptycirc & 
	\emptycirc & \fullcirc & \emptycirc \\
\cite{wang2020adaptive} & 
	\fullcirc & \halfcirc &  \emptycirc & \emptycirc & 
	\fullcirc & \emptycirc & \halfcirc  \\	
\cmidrule(l){1-1}\cmidrule(l){2-5}\cmidrule(l){6-8}

\cite{stricker2018reinforcement} & 
	\halfcirc & \fullcirc &  \emptycirc & \emptycirc & 
	\emptycirc & \emptycirc & \halfcirc \\
\cite{zhao2019improved} & 
	\fullcirc & \emptycirc &  \emptycirc & \emptycirc & 
	\emptycirc & \halfcirc & \fullcirc \\


\cmidrule(l){1-8}

\end{tabular}
\label{tab:reproducibility&validation_static}
\end{table} 


\subsection{Reproducibility}
\label{subsec:reproducibility}
\textbf{Setup and RL Design Clarity:} In the absence of standardisation, it becomes difficult not to forget to include all production setups details. 
In the literature at hand the presented setups were described transparently with respect to most of their details save for the exact problem size (e.g. number of machines in each work-center, total number of scheduled jobs) and sometimes some constraint details.
Similarly, more effort should be put into a clear presentation of all the RL design elements. 
Particularly the state and action designs suffer from a lack of clarity.

\textbf{Input Availability:} Ideally, in order to compare production scheduling approaches, the exact inputs, e.g. the job sequence with the operation precedence constraints and duration, should to be provided. 
Unfortunately, this is mostly not the case. 
Instead the input sampling scheme is given. 
Running experiments on randomly sampled inputs has the caveat of requiring a large number of experiments and statistical testing to ensure comparability with other approaches in the absence of reproducibility. Such an explicit statistical analysis is absent from the RL scheduling literature.
Moreover, the simulation code associated with the experiments should be provided as in \cite{kuhnle2020designing}.

\textbf{Reproducible Stochasticity:} Code availability alone does not suffice to ensure reproducible stochasticity, though it is a necessary condition. There is a simple way to ensure that a sequence of sampling actions produces the same output irrespective of the system the simulation is ran on: random number generator seeding (RNG). To ensure that the occurrence time of the stochastic events is independent of simulation control, one should sample \textit{all} the stochastic events together with their occurrence time during simulation initialization. These events can then be queued and triggered at the appropriate time. A simulation implementing this scheme is described in \cite{rinciog2021fabricatio}.

\subsection{Evaluation}
\label{subsec:baselines}
\textbf{Train-Test Split:} As with all machine learning approaches, RL schedulers are susceptible to overfitting. In deterministic setups, this refers to a situation where a strategy is developed  that is very well suited for the training instances, but cannot generalize well to unseen situations. 
Only reporting results on the production instances used for training is tantamount to assuming that the scheduling problem optimized is recurrent. 
In such recurrent, deterministic situations one may be better served by using other approaches, seen as whenever the comparison is available, RL fails to outperform the literature lower bounds, e.g. \cite{gabel2012distributed, martinez2011reinforcement, fonseca2018q}. 
In the dynamic setting, the test-train separation is not mandatory, since the stream of product specification induces a test-train separation, with the beginning of the stream being used for training and the latter part for testing.

\textbf{Sufficient Baselines:} RL solutions for stochastic settings are almost exclusively  baselined against simple heuristic solutions. It could be, however, that constructing schedules as if there were no stochasticity and simply ignoring plan deviations thereafter could be a competitive approach to RL or heuristics. Yet another approach would be to recompute a schedule on occurrence of unforeseen events. 
Constraint programming solvers such as ORTools \cite{ortools} allow for the use of a time limit (e.g. \cite{da2019industrial}) for the solution space search. Therefore, RL solvers for large production instances could at least be benchmarked against search, which is a standard OR strategy.

\textbf{Cherry Picking Potential:} Cherry picking as a concept, is very simple: One runs $n$ experiments, but only reports the results on $m < n$ of those that fit the own hypothesis. Whenever the results are reported on only a small number of experiments, the experiment design is not reproducible, and there is no statistical testing involved, this lingering suspicion remains. 

\section{Conclusion}
\label{sec:conclusion}
In this work we took steps towards standardizing RL experiments for production scheduling problems by introducing and extending an OR standard for the problem formulation (Section \ref{sec:setup}), describing the core elements of production scheduling RL design (Section \ref{sec:rl_modeling}), and proposing a scheme guaranteeing reproducibility and sufficient validation (Section \ref{sec:val}). 


The categories presented here were constructed to curtail three common problems of the RL scheduling literature. Firstly, possibly owing to the interdisciplinary nature of the field, the nomenclature used to describe scheduling setups is mostly inconsistent. This leads to a lack of clarity and redundant descriptive work, which in turn renders experiment reproducibility more difficult. Secondly, because of the large design space associated with RL production scheduling, necessary MDP details are sometimes imprecisely formulated. Thirdly, there is a validation gap for RL solutions to stochastic problems owing to a lack of experiment reproducibility and insufficient baselining. 


To consolidate the proposed standard, three further steps should be taken. To begin with, the more recently published RL scheduling literature should be 
indexed along the categories presented here. Additionally, the information used in the definition of states, actions and rewards should be exactly extracted and formalized. For instance, we should define individual features and rewards and not just their categories, as was done here. Finally, the standard should be used to implement or extend a benchmarking simulation such as the one in \cite{rinciog2021fabricatio}.

\section*{Acknowledgements}
\label{sec:ack}

This work was funded by the German Research Foundation through the Research Training Group 2193 \enquote{Adaption Intelligence of Factories in a Dynamic and Complex Environment}.

\printbibliography

\end{document}